\documentclass[conference]{IEEEtran}
\IEEEoverridecommandlockouts
\usepackage{cite}
\usepackage{amsmath,amssymb,amsfonts}
\usepackage{algorithm,algorithmic}
\usepackage{graphicx}
\usepackage{textcomp}
\usepackage{xcolor}
\def\BibTeX{{\rm B\kern-.05em{\sc i\kern-.025em b}\kern-.08em
    T\kern-.1667em\lower.7ex\hbox{E}\kern-.125emX}}
\usepackage[caption=false, font=footnotesize]{subfig}
\usepackage[export]{adjustbox}

\begin{document}

\title{PaaS: Planning as a Service for reactive driving in CARLA Leaderboard\\
}

\makeatletter
\newcommand{\linebreakand}{%
  \end{@IEEEauthorhalign}
  \hfill\mbox{}\par
  \mbox{}\hfill\begin{@IEEEauthorhalign}
}
\makeatother

\author{%
  \IEEEauthorblockN{%
    Nhat Hao Truong,
    Huu Thien Mai,
    Tuan Anh Tran,\\
    Minh Quang Tran,
    Duc Duy Nguyen,
    Ngoc Viet Phuong Pham 
  }%
}


\def\footnoterule{\relax%
    \kern -3pt
    \hrule width 2in
    \kern 2.6pt 
}

\maketitle
\begingroup\renewcommand\thefootnote{\textsection}
\endgroup

\newcommand{\frenet}{Fren\'et}

\newcommand{\tworowname}[2]{$\begin{array}{c}\textbf{ #1 } \\ \textbf{ #2 }\end{array}$ }

\newcommand{\multicolumnname}[2]{\multicolumn{1}{|p{1.2cm}|}{\centering \textbf{#1} \\ \textbf{#2}} }

\newcommand{\TC}{\mathit{TC}}

\begin{abstract}
End-to-end deep learning approaches have been proven to be efficient in autonomous driving and robotics. By using deep learning techniques for decision-making, those systems are often referred to as a black box, and the result is driven by data. In this paper, we propose PaaS (Planning as a Service), a vanilla module to generate local trajectory planning for autonomous driving in CARLA simulation. Our method is submitted in International CARLA Autonomous Driving Leaderboard (CADL), which is a platform to evaluate the driving proficiency of autonomous agents in realistic traffic scenarios. Our approach focuses on reactive planning in Frenet frame under complex urban street's constraints and driver's comfort. The planner generates a collection of feasible trajectories, leveraging heuristic cost functions with controllable driving style factor to choose the optimal-control path that satisfies safe traveling criteria. PaaS can provide sufficient solutions to handle well under challenging traffic situations in CADL. As the strict evaluation in CADL Map Track, our approach ranked 3rd out of 9 submissions regarding the measure of driving score. However, with the focus on minimizing the risk of maneuver and ensuring passenger safety, our figures corresponding to infraction penalties dominate the two leading submissions by 20 percent.

\end{abstract}


\section{Introduction}

\subsection{Motivation}
Autonomous driving is a rapidly growing field that has the potential to revolutionize transportation by providing safer, more efficient, and convenient transportation for people around the world. One of the most critical aspects of autonomous driving is motion planning, which involves determining the optimal path for a vehicle to follow to reach its destination while avoiding obstacles and adhering to traffic rules. Motion planning is a complex task that requires sophisticated algorithms, advanced sensors, and powerful computing resources. Furthermore, with the rise of autonomous vehicles, there is a growing need for robust and reliable algorithms that can safely and efficiently navigate complex driving scenarios. One of the key challenges in developing such algorithms is the lack of a standardized evaluation platform that can be used to compare the performance of different algorithms. The CARLA (Car Learning to Act) Leaderboard Challenge \cite{b35} is an open competition designed to address this challenge by providing a common platform for evaluating the performance of autonomous driving algorithms. The challenge is based on the CARLA simulator \cite{b36}, an open-source software platform that enables researchers and developers to test and develop their algorithms in a safe and controlled environment. In this paper, our main focus is on the motion planning stage in autonomous driving (see Section \ref{section-planning}). Moreover, we briefly introduce our approaches applied in CADL in Section \ref{section-carla-approach}. Finally, the evaluation of our result on CADL challenge compared to other competitors will be shown in Section \ref{section-eval}.

\subsection{Related work}
Motion planning is a crucial component of autonomous driving that involves generating a safe trajectory or path for the vehicle to follow in order to navigate through the environment. Many existing methods have been proposed to address this task, each with its own strengths and limitations.

Sampling-based methods randomly sample the environment and attempt to connect the samples, based on agent kinematics, to create a path, known as probabilistic sampling-based algorithms. Examples of these methods include Rapidly-exploring Random Trees (RRT) \cite{b6, b7} and Probabilistic Roadmaps (PRM) \cite{b8, b9}. Although the best path produced by these algorithms is often far from optimal, in \cite{b10}, the authors proposed the Rapidly-exploring Random Graphs (RRG) algorithm to ensure the returned solution is asymptotically optimal. The authors introduced RRT*, which is based on RRG to construct a tree from an existing graph, while also improving the cost-to-come value of the neighbor vertices for sampling points \cite{b11}. Although these methods are computationally efficient and can handle unknown environments, their performance suffers from random distribution, making them insufficient to apply in well-defined environments, such as urban or highway driving. To take advantage of structured environments, lattice planners are introduced in reactive motion planning. These methods produce a finite set of trajectories that sample over the spatio-temporal evolution space, satisfying agent kinematics, dynamics, and environmental constraints, such as obstacles, lane markings, and traffic signals. In \cite{b12, b13}, the authors decouple lateral and longitudinal movements to generate a dynamics profile under \frenet{} frame to generate a set of trajectories, defined according to the driving state of the agent. Following the generation of the trajectory set, obstacle representation is a crucial aspect of ensuring safe and efficient navigation. Obstacle representations, such as circular \cite{b14}, triangular \cite{b15}, and rectangular \cite{b16}, are proposed to improve coverage and computational efficiency.

Optimization-based methods use optimization techniques to find the optimal path that satisfies a set of constraints, such as minimum time or energy consumption. Model Predictive Control (MPC) \cite{b16, b19, b20} and Differential Dynamic Programming \cite{b21, b22} are examples of optimization-based methods. In urban driving scenarios, the automated vehicle needs to solve a joint optimization of neighboring vehicles (NV) costs over a receding horizon. To handle such task, a mixed integer quadratic programming (MIQP) formulation of the MPC is presented in \cite{b17} to handle the multi-input multi-output (MIMO) control problem with indicator variables and disjunctive constraint. Based on this idea, \cite{b18} suggests adaptive interactive mixed integer MPC (aiMPC), which captures dynamics and collision avoidance constraints to predict the trajectory of the ego and NV in MPC horizon.

AI-based methods are another category of motion planning algorithms that use machine learning, deep learning, or reinforcement learning algorithms to learn mappings between the current state of the vehicle, sensors, environments, and the optimal action to take, namely end-to-end autonomous driving. These methods can handle complex and dynamic environments, but they require large amounts of data and time to train. \cite{b23} vectorizes the sensors' information with multiple modalities, such as OpenDRIVE HD Map, Radar, LiDAR, and front camera image, into same-sized inputs. With different sensor domains, CNN-based fusion layers, or Transformer-based ones \cite{b24, b28, b31}, are proposed to offer a chance for these sensor data to find relations with each other. \cite{b25, b27} offer deep reinforcement learning (DRL) techniques that learn state-action policies from offline replay buffer (RB) (recorded from expert data) and online exploration agent from online RB.

\section{MOTION PLANNING METHOD}\label{section-planning}

In this section, the reactive motion planning task is formulated in \frenet{} frame by the decoupling of lateral and longitudinal movements to generate dynamics profiles for the agent. Furthermore, the constraints in urban driving scenarios are explained in detail.

\subsection{Trajectory Planning in \frenet{} Frame} \label{section-trajectory-planning}
The work in this section is based on the proposal of using \frenet{} Frame method in \cite{b12}. Following this paper, we generate the trajectory sets using multiple terminal conditions to follow the center lines on the road (reference lines), then the conversion from \frenet{} frame to Cartesian frame is performed to spatio-temporal collision checking with other surrounding agents. As the most advantage of using \frenet{} frame in the trajectory generation stage is that the reference lines are standardized in the same form of a straight line with the difference in road boundaries. Nevertheless, the side effects of the usage of this method are mentioned in \cite{b37}. To avoid those disadvantages, the sampling of trajectory set is executed in \frenet{} frame using a simplified kinematic model and driving comfort constraints, then the valid set is evaluated with environment constraints (detailed in Section \ref{section-constraints}) in Cartesian space. 

The \frenet{} frame is composed of the tangential and normal vector to model the reference curve in Cartesian coordinate. In motion planning using \frenet{} frame, the tracking problem is composed of two factors: lateral and longitudinal offset along the temporal dimension, namely $d(t)$ and $s(t)$ respectively. The movement profiles of our agent are defined in lateral direction $D(t)=[d(t),\dot{d}(t),\ddot{d}(t), \dddot{d}(t)]$ and longitudinal one $S(t)=[s(t),\dot{s}(t),\ddot{s}(t),\dddot{s}(t)]$. We use quintic polynomials for the generation of $D(t)$ in lateral space and $S(t)$ in longitudinal space within time interval $T$. 

The generated trajectory needs to satisfy the jerk-optimal constraints. Therefore, the total jerk is the accumulation of jerk over the planning horizon, applied in both lateral jerk polynomial $\dddot{s}(t)$ and longitudinal jerk  $\dddot{d}(t)$

\begin{equation}
J_s = \int_{0}^{T} \dddot{s}^2(t) dt; \text{ } J_d = \int_{0}^{T} \dddot{d}^2(t) dt.
\end{equation}

Furthermore, to satisfy the comfort movement, the trajectory must obey the conditions that 
\begin{equation}
\dddot{s}(t) < J_{max}, \dddot{d}(t) < J_{max}, \forall t \in [0, T],
\end{equation}
where $J_{max}$ is the maximum comfort jerk.

In urban driving, many modes are selected and executed while traveling to ensure the continuity of movement. These modes, in descending order of aggressiveness, are merging, following, velocity-keeping in free space, and stopping. The behavior layer evaluates the consequences of mentioned modes to make decisions. In the scope of this paper, we use solely the total cost function to select the most optimal trajectory and the driving style of our agent is configurable by parameters regarding these functions. 

Each mode has two cost functions related to lateral and longitudinal space in \frenet{} frame. The start condition is the current state of the ego agent, which is retrieved from sensors (explained in Section \ref{section-localization}). The longitudinal and lateral terminal condition is given on the selected mode, namely $\TC$. The evaluation of trajectory corresponding to terminal condition is formulated by cost function $C$.

We define the common terminal state and cost function for longitudinal trajectory generation as

\begin{equation}\label{eq-terminal-longi}
\TC_{ij} = [s_j, \dot{s}_j, \ddot{s}_j, T_{i}],
\end{equation}
\begin{equation}
C_s = k_j J_s + k_t T_i + k_s(s_d - s_j)^2,
\end{equation}
with $s_d$ is the target distance in the longitudinal frame.

In the stopping mode, the agent has to stop its movements before the longitudinal position of the traffic sign (or red light), which is $s_d$. Following \eqref{eq-terminal-longi}, the terminal condition with the longitudinal difference $\Delta {s_j}$ is 

\begin{equation}
\TC_{ij} = [s_j, \dot{s}_j, \ddot{s}_j, T_{i}] = [s_{d} - \Delta {s_j} , 0, 0, T_{i}].
\end{equation}

In following mode or cruise control, we maintain a safe distance from the preceding vehicle. The state corresponding to the preceding vehicle $S_{pv} = [s_{pv}(t), \dot{s}_{pv}(t), \ddot{s}_{pv}(t)]$. We define the predicted terminal state of the target vehicle in the prediction horizon as $\hat{S}_{pv}(T) = [\hat{s}_{pv}(T), \hat{\dot{s}}_{pv}(T), \hat{\ddot{s}}_{pv}(T)]$, which is retrieved from the Trajectory Prediction module (detailed in Section \ref{section-trajpred}). Applying into \eqref{eq-terminal-longi}, the desired states of our agent are as follows

\begin{equation}
\begin{aligned}
& s_{j} = \hat{s}_{pv}(T) -\left[D_0+\tau \hat{\dot{s}}_{pv}(T)\right], \\
& \dot{s}_{j} =\hat{\dot{s}}_{pv}(T) \pm \Delta{\dot{s}_j} -\tau \hat{\ddot{s}}_{pv}(T), \\
& \ddot{s}_{j} =\hat{\ddot{s}}_{pv}(T),
\end{aligned}
\end{equation}
with $D_0$ is the constant safety distance with the other agent, followed by the gap defined by constant time-to-collision $\tau$.

In merging mode, it requires the agent to keep an appropriate distance from both preceding and following vehicles. Using the predicted trajectories of preceding $\hat{S}_{pv}(t)$ and following $\hat{S}_{fv}(t)$ vehicles, we can define the target point in the terminal condition \eqref{eq-terminal-longi} as

\begin{equation}
s_{j}=\frac{1}{2}\left[\hat{s}_{pv}(T)+ \hat{s}_{fv}(T)\right].
\end{equation}


Corresponding to velocity-keeping mode, we keep our agent to maintain its velocity without specific $s_d$. Therefore, we use quartic polynomials for the generation of $S(t)$ as shown in Fig. \ref{fig-longitudinal}. The terminal state, without target longitudinal distance, is formulated as follows, where $\dot{s}_d$ is the desired velocity
\begin{equation}
\TC_{ij}= [\dot{s}_{d} \pm \Delta{\dot{s}_j}, 0, T_{i}],
\end{equation}
\begin{equation}
C_v = k_j J_s + k_t T_i + k_s(\dot{s}_d - \dot{s}_j)^2.
\end{equation}

\begin{figure}[htbp]
    \centering
  \subfloat[Longitudinal velocity]{%
       \includegraphics[height=0.5\linewidth]{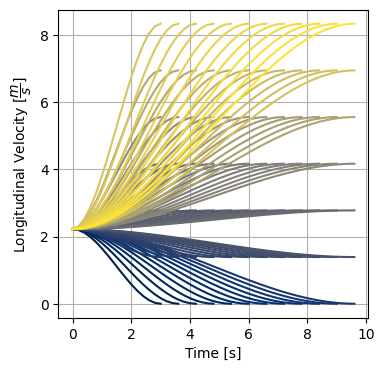}}
  \subfloat[Longitudinal distance]{%
        \includegraphics[height=0.5\linewidth]{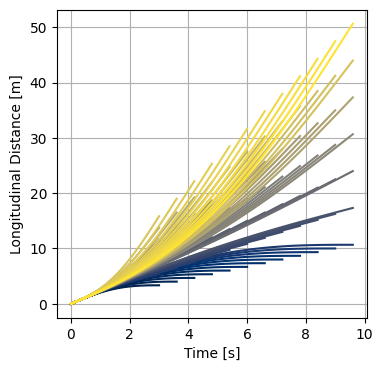}}
    \hspace{0.5cm}

  \caption{The longitudinal profile based on velocity tracking is shown as (a) longitudinal velocity, (b) corresponding longitudinal distance. Each solution is color-mapped by its longitudinal target velocity $\dot{s}_j$.  $\Delta v = 5 km/h = 1.38 m/s$ and $\Delta T = 0.2s$}
  \label{fig-longitudinal} 
\end{figure}

Regarding lateral space, we generate the trajectory set in lateral space with multiple offset $d_j$ (with the difference $\Delta d$) and time interval $T_i$ in the terminal condition. The final state is when our agent successfully tracks the reference line ($d_T = 0$). The movement profiles of lateral movement are shown in Fig. \ref{fig-lateral}. Similar to longitudinal space, the terminal condition is defined as 
\begin{equation}
\TC_{ij}=[d_j, 0, 0, T_{i}],
\end{equation}
\begin{equation}
C_d = k_j J_d + k_t T_i + k_s d_j^2.
\end{equation}

\begin{figure}[htbp]
    \centering
  \subfloat[Lateral movement]{%
       \includegraphics[height=0.5\linewidth]{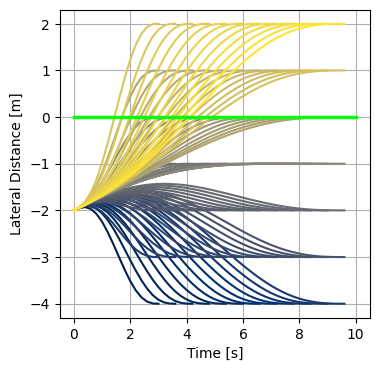}}
  \subfloat[Lateral velocity]{%
        \includegraphics[height=0.5\linewidth]{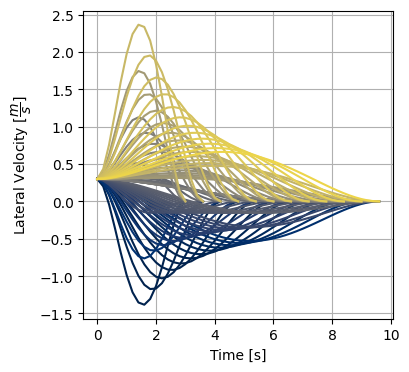}}
        
  \caption{Illustrations of (a) lateral movement, (b) corresponding lateral velocity. The green line is the target lateral offset with the reference line (at zero). Multiple solutions in (a) are generated with multiple lateral and time interval differences, with $\Delta d = 1m$ and $\Delta T = 0.2s$ respectively, color-mapped by its lateral offset $d_j$.}
  \label{fig-lateral} 
\end{figure}

\subsection{Environment Constraints}\label{section-constraints}

In our urban driving task, the constraints include the collision boundaries of NV and pedestrians, road barriers, traffic lights, and road signs.

First of all, in CARLA simulation, traffic lights and road signs' positions can be retrieved from the provided map. However, the state of the traffic light (green, yellow, red) can only be identified visually by the cameras (by methods explained in Section \ref{section-perception}). We convert the queried positions on the global frame and their states as terminal conditions in \frenet{} frame in the motion planning module, as described in the previous section.

Furthermore, we execute collision checking on the sampling set of trajectories on Cartesian space. Each trajectory is denoted by $\chi = [x(t), y(t), \theta(t), v(t), a(t)]$, where $[x(t), y(t)]$ denotes the position, $\theta(t)$ represents the orientation angle, $v(t)$ indicates the velocity and $a(t)$ corresponds the acceleration of trajectory points with reference to (w.r.t) our ego agent. Through collision checking, we assess a set of feasible trajectories denoted as $\Phi$, resulting in the formation of a collision-free set known as $\Phi^*$.

In order to enhance the computational efficiency of collision checking, we adopt a circle representation that encompasses the rectangular shape of the vehicles, including our agent. This approach involves employing an odd number of circles, denoted as $n$, to cover the overall shape. As illustrated in Fig. \ref{fig-disk-model}, there are two types of representation that are being used for collision checking: three-disk and five-disk models. While both models are able to cover the whole rectangular section, there is a trade-off between computational efficiency by using the three-disk model and precise representation by using the latter model, which has smaller circle coverage by 9.02\%. Based on our experiment, we decide to use the three-disk model to have the computational advantage, with time complexity being $\mathcal{O}(n^n)$. Each disk in the model could be constructed by 2D center point and radius by $D = [x, y, R]$ in the relative coordinate of the vehicle. Therefore, we define that
\begin{equation}
D_{1} =\left[0, 0, R_{max} \right],
D_{2,3} = \left[\pm \frac{l}{3}, 0 , R_{max} \right],
\end{equation}
where $D_{1}$,$D_{2}$,$D_{3}$ are the models of center, front, and back disks respectively. $w$ is width, $l$ is length of the vehicle, and $R_{max} = \left(\frac{w^2}{4} + \frac{l^2}{36} \right)$.

Within 2D space, the collision is violated when

\begin{equation}
\label{eqn:circle_check}
  ||p_{ego}^i - p_{nv}^j|| <  R_{ego} + R_{nv}, i,j = 1,\ldots,3,
\end{equation}
with $p_{ego}^i$ is the center location of ith disk of ego vehicle and $p_{nv}^j$ is the center location of jth disk of NV. $R_{ego}, R_{nv}$ are disk radiuses of ego vehicle and NV respectively.

\begin{figure}[htbp]
    \centering
  \subfloat[Three-disk model]{%
       \includegraphics[width=0.28\linewidth]{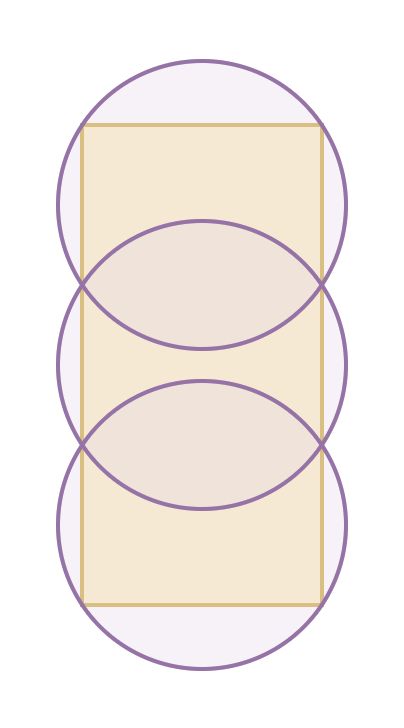}}
    \hspace{1.2cm}
  \subfloat[Five-disk model]{%
        \includegraphics[width=0.28\linewidth]{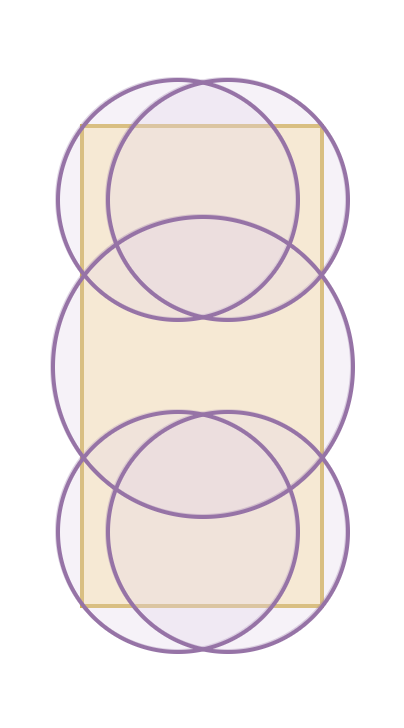}}
  \caption{Disk representation (purple) of the rectangular vehicle (orange). By using more disks, the rectangular could be represented more precisely. However, there will be computational overhead in the collision-checking process.}
  \label{fig-disk-model} 
\end{figure}

In real scenarios, the collision needs to be checked in three dimensions, including 2D space and additional dimension of time, within the planning time horizon $t_h$. At every future timestep $t_k$ over $t_h$, \eqref{eqn:circle_check} must not be satisfied to consider a trajectory collision-free. The formulation of collision violation, in this case, is defined as

\begin{equation}
||p_{ego}^{i, t_k} - p_{nv}^{j, t_k}|| <  R_{ego} + R_{nv}, \forall t_k \in [0, t_h], i,j = 1,\ldots,3.
\end{equation}

Following that, by utilizing the collision-free set $\Phi^*$, we ensure the executability of all feasible trajectories for the ego vehicle. Next, we determine the optimal solution $\chi^*$ by considering the trajectory with the lowest cost value. It is important to note that in this particular case, the term "optimal" does not refer to the concept of optimality in trajectory optimization problems. Instead, it signifies the selection of the most favorable feasible trajectory from the set of sampled trajectories. By continually replanning the optimal trajectory within a fixed time interval, the module guarantees consistency in the resulting trajectory over time.

\section{CARLA LEADERBOARD APPROACH}\label{section-carla-approach}

In CADL, there are two tracks to submit the solution, namely SENSORS and MAPS tracks. In SENSORS, the set of sensors provided includes GNSS, IMU, LiDAR, RADAR, RGB camera (limit to 4 units), and Speedometer. Similarly, the MAPS track provides the same set with an additional OpenDRIVE map. Our approach is tested and submitted on MAPS track. In this section, because of the limited scope of this paper, we briefly introduce the methods, without digging into details, that we use to handle the sensor signals.

The flow of PaaS, which includes the Motion Planning module in Section \ref{section-planning}, as well as the Localization, Perception, and Trajectory Prediction modules described in this section, is presented in Algo. \ref{alg_paas}. The initial steps involve extracting semantic information about the surrounding environment (Line 1 to 4). This extracted information is then utilized in the motion planning modules, resulting in the generation of a set of feasible and collision-free trajectories (Line 5 to 6). Finally, each trajectory is evaluated using the cost function, leading to the selection of the optimal trajectory (Line 7).

 \renewcommand{\algorithmiccomment}[1]{//#1}

 \begin{algorithm}[H]
 \caption{Process of PaaS in CADL}
 \begin{algorithmic}[1]
  \label{alg_paas}

 \renewcommand{\algorithmicrequire}{\textbf{Inputs:}}
  \renewcommand{\algorithmicensure}{\textbf{Output:}}

    \REQUIRE Navigational sensor signals $\Psi$; Camera images $I$; Point cloud $PC$; Reference path $\xi$
    \ENSURE $\chi^{*}$ \hfill\COMMENT{optimal trajectory}
    \STATE  $S_{mo} \leftarrow$ MovingObjectDetection($PC$)
    \STATE  $Z_{ego} \leftarrow$ TrafficSignDetection($I$)
    \STATE  $\hat{X} \leftarrow$ Localization($\Psi$)
    \STATE  $\hat{\Pi} \leftarrow$ TrajectoryPrediction($S_{mo}$) 
    \STATE  $\Phi  \leftarrow$ MotionPlanning($\xi, \hat{X}, Z_{ego}$) 
    \STATE  $\Phi^{*} \leftarrow$ CollisionCheck($\Phi, \hat{\Pi}$) 
    \STATE  $\chi^{*} \leftarrow $ min($\Phi^{*}$) \hfill\COMMENT{select path with lowest cost}
    
 \end{algorithmic} 
 \end{algorithm}

\subsection{Localization}\label{section-localization}
To retrieve an estimated state of ego vehicle 
\begin{equation} 
\hat{X}=[\hat{x}, \hat{y}, \hat{\theta}, \hat{v}, \hat{a}],
\end{equation}
the method \cite{b39} relies on Kalman Filtering (KF), which uses navigational sensor signals
\begin{equation}
\Psi = [\lambda, \phi, \varphi, va_x, va_y, va_z, al_x, al_y, al_z, \psi],
\end{equation}
where latitude $\lambda$, longitude $\phi$, altitude $\varphi$ are obtained from GNSS information, angular velocity $[va_x, va_y, va_z]$, linear acceleration $[al_x, al_y, al_z]$, yaw compass $\psi$ are derived from IMU sensor.

\subsection{Perception} \label{section-perception}

First of all, we use the point cloud data $\mathit{PC}$ from the LiDAR sensor to handle the detection of surrounding vehicles and pedestrians. The whole process is described as Moving Object Detection. We apply a pre-trained model from \cite{b40} to predict the bounding boxes of surrounding vehicles and pedestrians, called moving obstacles. The detection outputs are objects' 3D shapes $[h_{mo}, w_{mo}, l_{mo}]$, 3D positions $[x_{mo}, y_{mo}, z_{mo}]$, yaw angles $\psi_{mo}$, and the detection confidences. Subsequently, we apply the KF to track and predict velocity $v_{mo}$, acceleration $a_{mo}$ of the moving obstacles. The state that describes a moving obstacle is
\begin{equation}
S_{mo} = [x_{mo}, y_{mo}, \psi_{mo}, v_{mo}, a_{mo}].
\end{equation}

On the other hand, the detection and recognition of traffic lights, and traffic signs are executed from the front camera's images, following the detection model in \cite{b41}. Firstly, we collect the training images from our test run in CARLA simulator. Secondly, we re-train the model using the approach of transfer learning. Eventually, with the output of the trained model, which are 2D location including coordinates and dimensions $P_{im} = [x, y, w, h]$ on image and the recognition of traffic light state, traffic sign $S_{sign}$, we are able to map $P_{im}$ in image coordinate into $P_{ego}$, which is the relative position w.r.t ego vehicle's coordinate,  using simple kinematic transformation and OpenDRIVE map. The final output state of the traffic light/sign w.r.t the ego vehicle will be 
\begin{equation}
Z_{ego} = [P_{ego}, S_{sign}].
\end{equation}

Furthermore, to improve efficiency, we extend the model to detect junctions that contain traffic lights to avoid false positive detections. Illustrated in Fig. \ref{detect-traffic-light}, our autonomous vehicle advances towards the intersection, identified by the blue bounding box and in conjunction with the detected traffic light. Incorporating this comprehensive information, the algorithm effectively executes trajectory planning while ensuring compliance with traffic regulations.

\begin{figure*}[htbp]
\centering
\subfloat[]{%
\adjincludegraphics[width=5cm,trim={0 0 {.3\width} 0},clip]{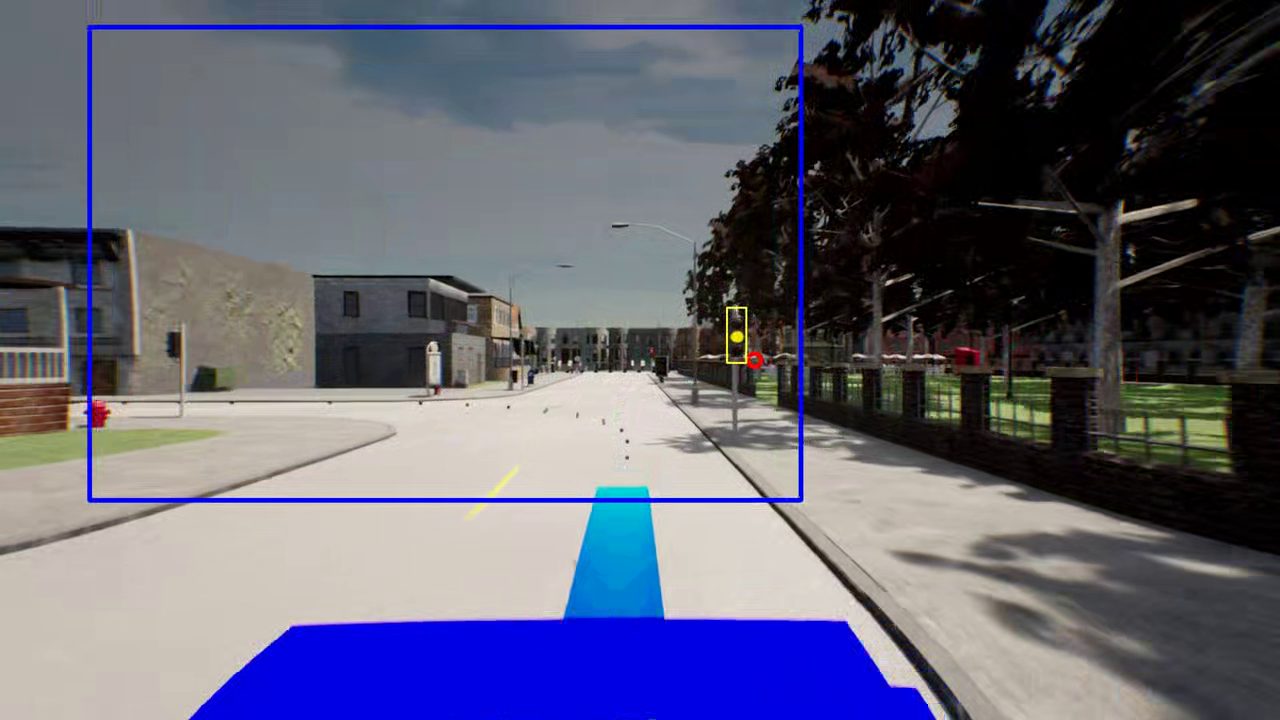}}
\hspace{0.2cm}
\subfloat[]{%
\adjincludegraphics[width=5cm,trim={0 0 {.3\width} 0},clip]
{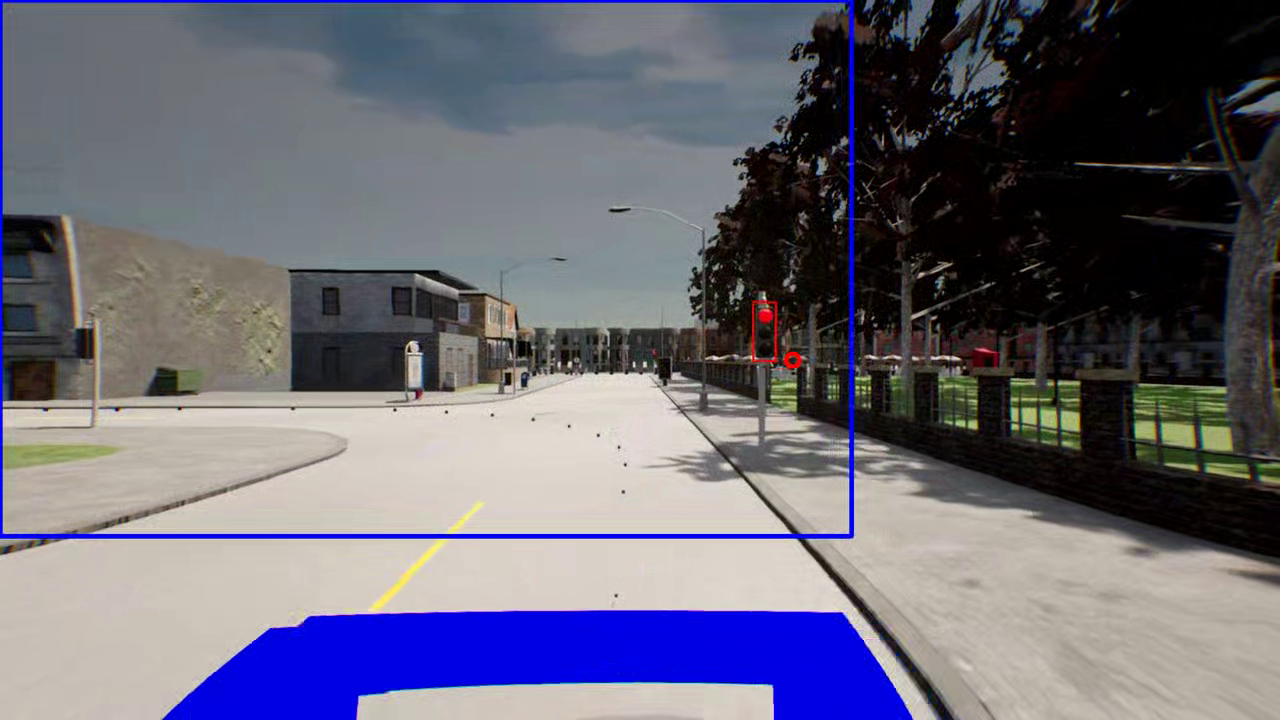}}
\hspace{0.2cm}
\subfloat[]{%
\adjincludegraphics[width=5cm,trim={0 0 {.3\width} 0},clip]{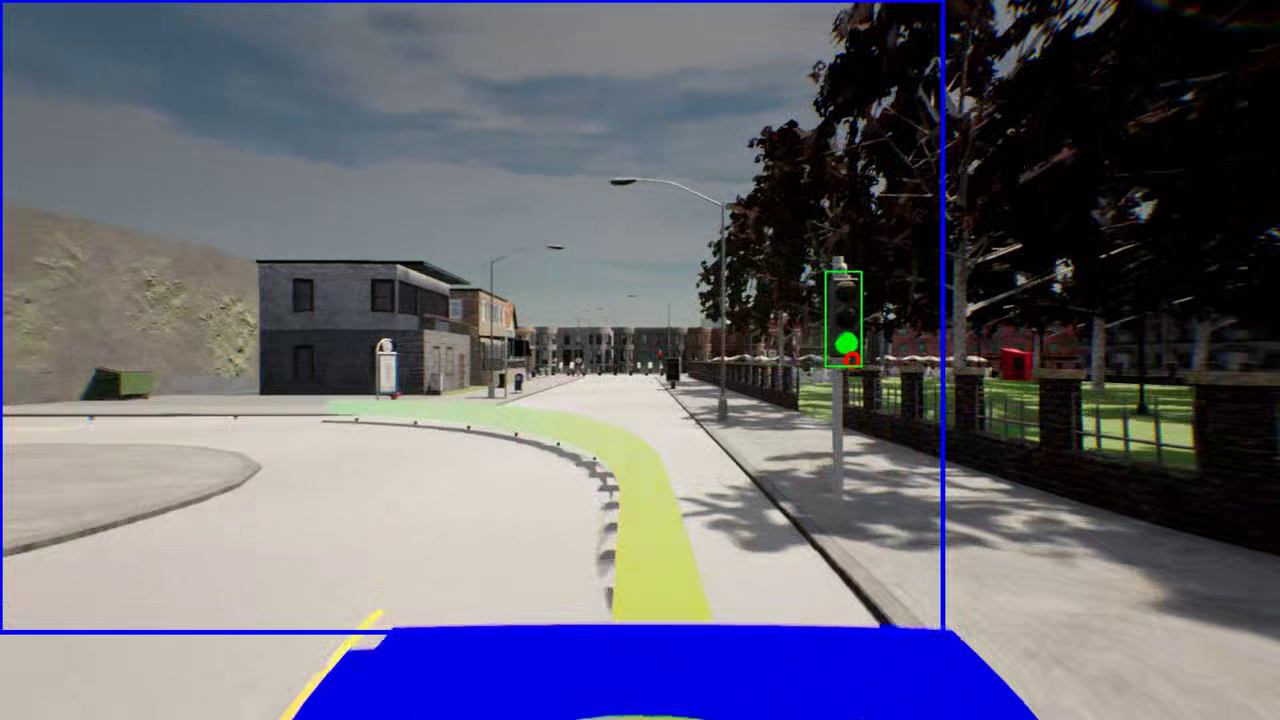}}

\caption{Illustrations of traffic light detections captured from our vehicle's front camera. The vehicle encounters different states as it approaches the traffic light: (a) yellow - the planned trajectory is stopped before the pole; (b) red - no planning is executed; (c) green - the planned trajectory making a left turn. The approximate traffic light positions are mapped from the OpenDRIVE map, shown as a red circle in the images.}

\label{detect-traffic-light}
\end{figure*}

\subsection{Trajectory Prediction}\label{section-trajpred}
Based on the dynamic state of the moving obstacle $S_{mo}$ and OpenDRIVE map given in CARLA simulator, we can predict its future trajectory in the planning horizon.

The original map topology in CARLA contains the tuple of pairs of waypoints located either at the beginning or end point of a road. Because each road has a different length, we create a dense graph to represent the given topology with a fixed sampling distance. An R-Tree is additionally used for graph index querying in spatial dimensions. At first, we assume that other vehicles in the simulation strictly follow the road center without any deviation. Based on the target vehicle's position, the corresponding node in the dense graph is queried using R-Tree indexing. Then, the Breadth-First Search algorithm is applied to the dense graph to extract the possible paths that have their length closed to a desired distance, which is approximated by the current vehicle dynamic states. Finally, a set of predicted trajectories $\hat{\Pi}$ is generated using the Pure Pursuit algorithm.

\section{RESULT AND DICUSSION}  \label{section-eval}

\subsection{Metrics}

In the CADL challenge, a set of metrics is provided to describe the driving proficiency of an agent. The main metrics are as follows
\begin{itemize} 
\item \textbf{Route completion} is the percentage of the route distance completed by an agent. The evaluation stops for each scenario when the violation occurs, such as collision and agent being blocked for a specific interval (180 seconds).

\item \textbf{Infraction penalty} is the metric used to show how safe the maneuver is. The value will decay for each time the agent commits a violation, based on the corresponding ratio. This metric value is in the range of $[0, 1]$.

\item \textbf{Driving score} is the main metric of the CADL, which is used to rank the participants. It is the product between the route completion ratio and the infraction score. Unit is $\%$.

\end{itemize}

In CADL, there are several types of infractions, such as Collisions with pedestrians, Collisions with vehicles, Collisions with layout, Running a red light, Route deviation, and Agent blocked. Each type of infraction has a specific penalty coefficient that impacts the overall score of the infraction penalty.

\subsection{Evaluation / Result on CADL}

When evaluating the results of our approach on CADL, we compared our method with other participants in the MAPS track of the CADL challenge \cite{b35}. The results of running test set provided by CADL are shown in Table \ref{table-cadl}. Our PaaS ranks third on the leaderboard, achieving a driving score of 48.24. However, our approach has the highest infraction penalty of 0.84, indicating that our agent can navigate safely throughout the given scenarios in the CADL challenge.

By examining the details of infractions and the behavior of our ego agent in typical scenarios depicted in Fig. \ref{scenarios}, we can conclude that our agent successfully avoids the violations encountered in CADL. Regarding infractions involving dynamic or surrounding agents, our figure for Collision with vehicles ranks first. This clearly demonstrates that our agent is capable of performing safe maneuvers. For instance, in Fig. \ref{stop_and_maintain}, the preceding vehicle stops at a red light, and our ego agent plans a trajectory to stop and maintain a safe distance from it. Additionally, in Fig. \ref{pass_pedestrian}, our agent effectively navigates through a junction with many pedestrians by incorporating spatial-temporal trajectory prediction, while maintaining good tracking performance with the reference path.

Another notable example is shown in Fig. \ref{escape_junction}, where our ego agent confidently executes a right turn at a junction, knowing that the nearby vehicle will not obstruct its path. Similarly in Fig. \ref{slow_speed_pedestrian}, the ego agent considers the predicted trajectory of a crossing pedestrian but still makes a prudent decision by maintaining a low speed for forward acceleration, ultimately avoiding collisions with moving obstacles.

In terms of adhering to traffic rules, our agent ranks second in red light infractions, which is primarily influenced by the performance of the Perception module. Moreover, we achieve perfect scores in Route deviations and Collision layout metrics, indicating a decent tracking performance with the reference path.

\begin{figure*}[htbp]

\centering
\subfloat[\label{stop_and_maintain}]{%
\adjincludegraphics[width=4.2cm,trim={{.3\width} {.2\width} {.3\width} {0\width}},clip]{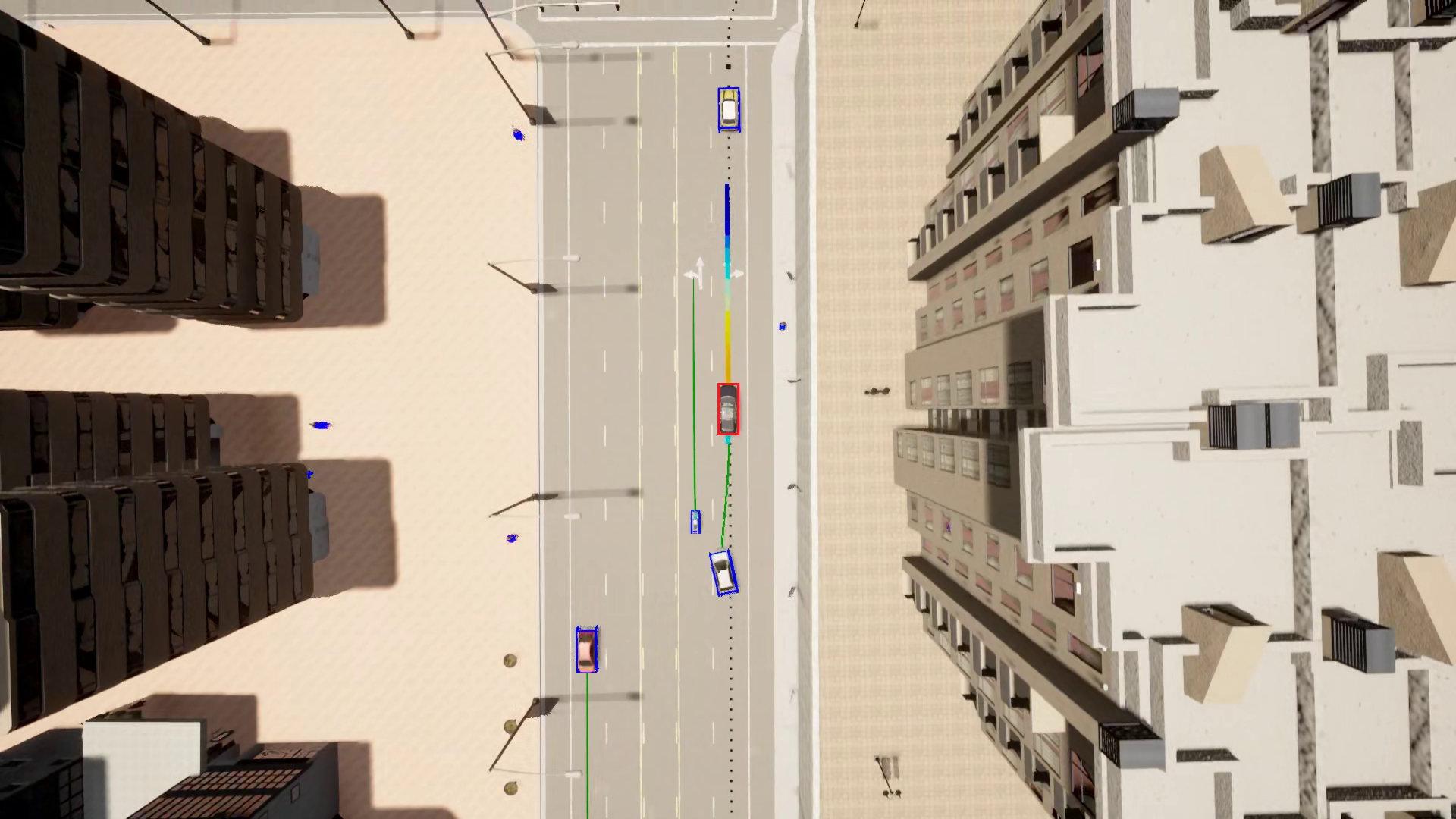}}
\hspace{0.2cm}
\subfloat[\label{pass_pedestrian}]{%
\adjincludegraphics[width=4.2cm,trim={{.3\width} {.2\width} {.3\width} {0\width}},clip]
{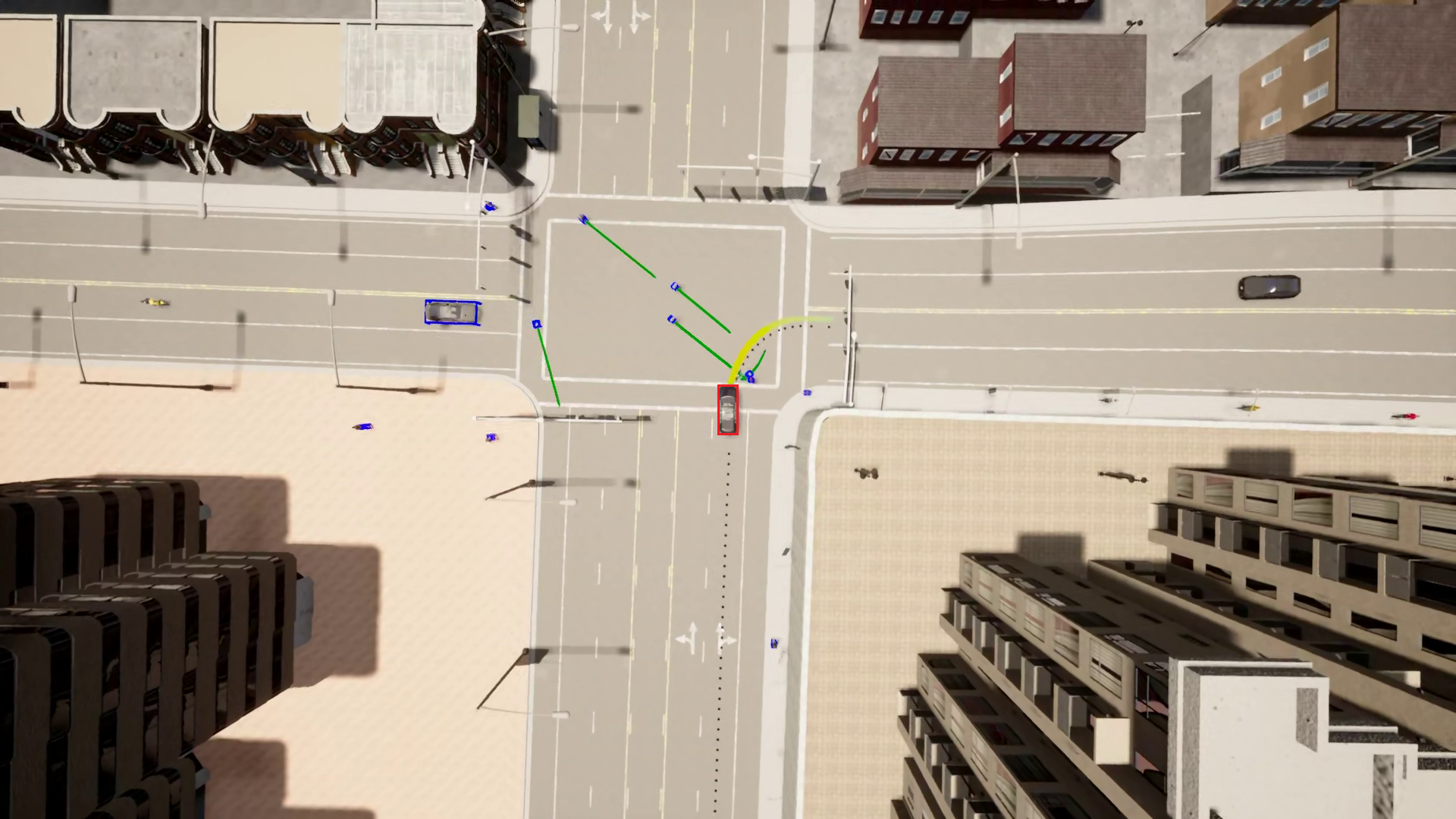}}
\hspace{0.2cm}
\subfloat[\label{escape_junction}]{%
\adjincludegraphics[width=4.2cm,trim={{.3\width} {.2\width} {.3\width} {0\width}},clip]{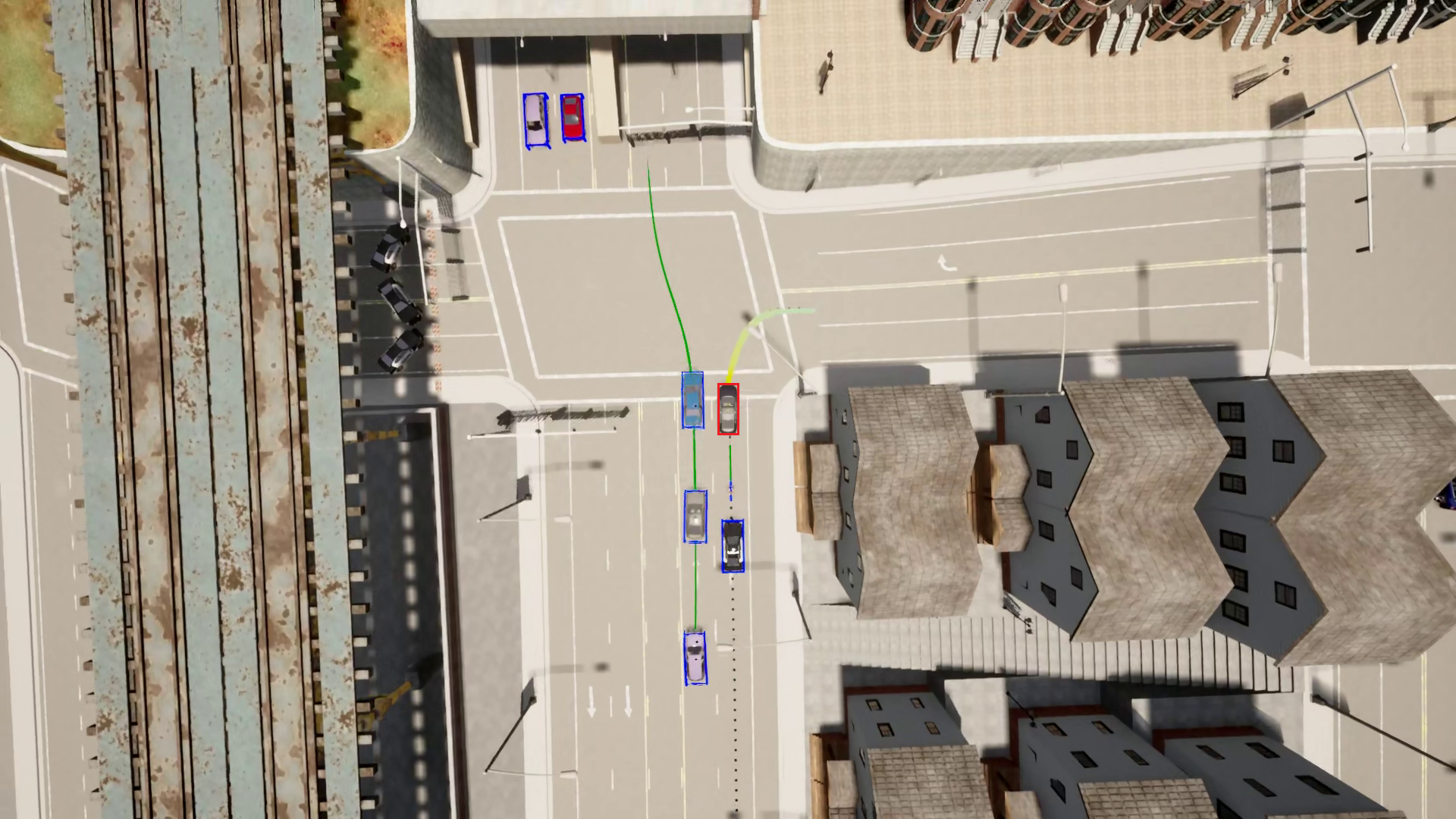}}
\hspace{0.2cm}
\subfloat[\label{slow_speed_pedestrian}]{%
\adjincludegraphics[width=4.2cm,trim={{.3\width} {.2\width} {.3\width} {0\width}},clip]{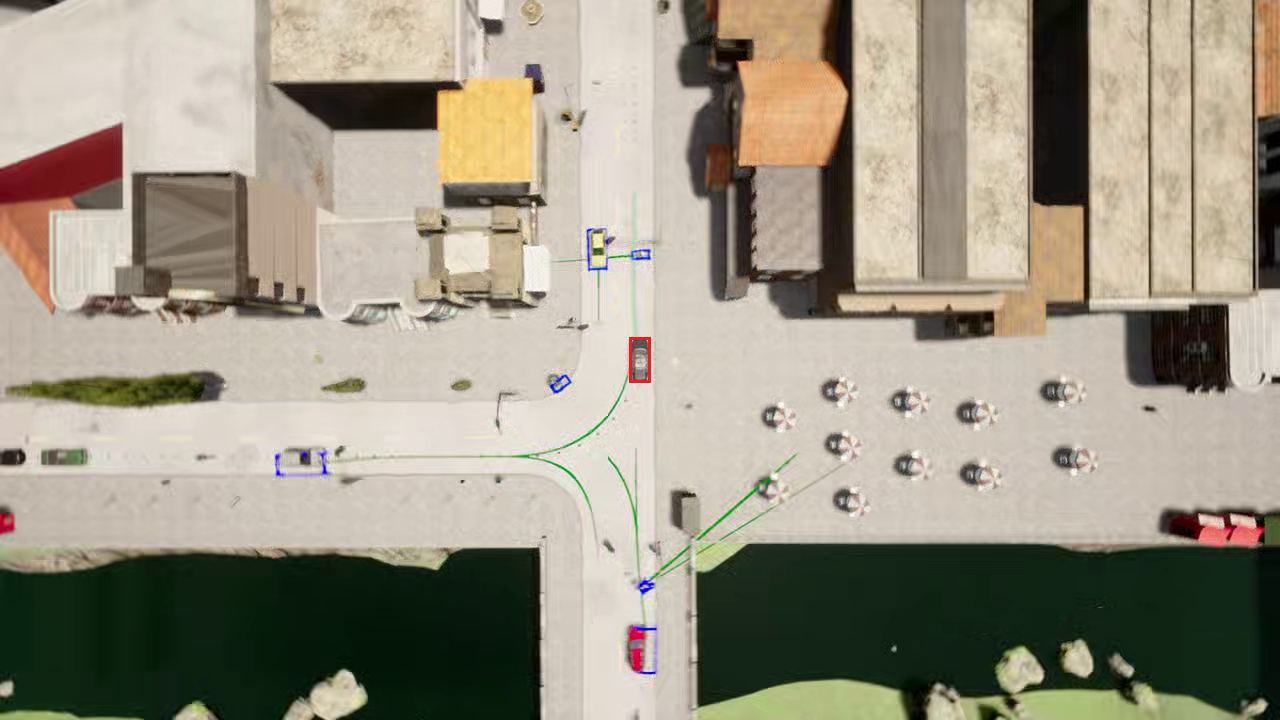}}

\caption{Illustrations of our ego vehicle maneuvering through typical scenarios. The Localization module determines the position of our system relative to the map. The Perception module identifies moving obstacles, such as vehicles and pedestrians, within the LiDAR's field of view and represents them with blue bounding boxes. The green paths illustrate the predicted trajectories of these obstacles. The reference path is visualized as a black dotted line. Using this information, the Motion Planning module generates the optimal trajectory for our agent, which is color-coded based on velocity: blue represents stopping speed, orange indicates low speed, and red represents high speed.}
\label{scenarios}
\end{figure*}

However, the number of pedestrian collisions in our approach is higher compared to the participants ranking first and second. This can be attributed to a few factors. Firstly, the effectiveness of LiDAR in detecting pedestrians is limited due to the small size of humans relative to the sensor's resolution. Secondly, the predicted trajectories of pedestrians do not always align with their actual behavior in the CADL environment. Furthermore, there are instances in CADL where pedestrians unexpectedly cross the road from the blind spots of our agent, and we were unable to react promptly to such behavior. Unfortunately, the testing scenarios only provide metrics and do not include visual images from cameras, making it more challenging to analyze the root cause.

As discussed in Section \ref{section-trajectory-planning}, our approach relies solely on the cost function of the trajectory to determine the most optimal path. However, in certain specific scenarios, selecting the trajectory with the lowest cost does not always lead to the best overall solution. This phenomenon, known as local minima, is particularly prevalent in dense traffic situations. Consequently, our figure of Agent blocked infractions is unusually high compared to the performance of other participants.

\begin{table*}[htbp]
\caption{Test-set evaluation result on CARLA Leaderboard Challenge}
\begin{center}

\begin{tabular}{|c|c|c|c|c|c|c|c|c|c|c|c|c}
\hline  
\textbf{Rank} & 
\textbf{Name} & 
\multicolumnname{Driving}{score}&
\multicolumnname{Route}{compl.}&
\multicolumnname{Infrac.}{penalty}&
\multicolumnname{Collision}{pedes.}&
\multicolumnname{Collision}{vehicle}&
\multicolumnname{Collision}{layout}&
\multicolumnname{Red light}{infrac.}&
\multicolumnname{Route}{deviations}&
\multicolumnname{Agent}{blocked}
\\
\hline
&
&
$\%, \uparrow$ &
$\%, \uparrow$ &
$[0, 1], \uparrow$ &
$\# / \mathrm{km}, \downarrow$ &
$\# / \mathrm{km}, \downarrow$ & 
$\# / \mathrm{km}, \downarrow$ & 
$\# / \mathrm{km}, \downarrow$ & 
$\# / \mathrm{km}, \downarrow$ & 
$\# / \mathrm{km}, \downarrow$
\\
\hline
1&
Map TF++&  
\textbf{61.17}&81.81&0.70&\textbf{0.01}&0.99&0.00&\textbf{0.08}&0.00&0.55
\\
2&
MMFN+ \cite{b23}&  
59.85&\textbf{82.81}&0.71&0.01&0.59&0.00&0.51&0.00&\textbf{0.06}
\\
3&
\textbf{PaaS (ours)}&
48.24&60.68&\textbf{0.84}&0.10&\textbf{0.23}&\textbf{0.00}&0.13&\textbf{0.00}&4.13
\\
4&
GRI-based DRL \cite{b25}&
33.78&57.44&0.57&0.00&3.36&0.50&0.52&1.47&0.80
\\
5&
MMFN \cite{b23}&
22.80&47.22&0.63&0.09&0.67&0.05&1.07&0.00&1003.88
\\
6&
Techs4AgeCar+&
18.75&75.11&0.28&1.52&2.37&1.27&1.22&0.17&1.28
\\
7&
Pylot&
16.70&48.63&0.50&1.18&0.79&0.01&0.95&0.44&3.30
\\
8&
CaRINA \cite{b38}&
15.55&40.63&0.47&1.06&3.35&1.79&0.28&0.34&7.26
\\
9&
Techs4AgeCar&
12.63&61.59&0.33&2.25&0.63&0.00&0.96&0.02&1.34
\\
\hline
\multicolumn{2}{l}{$\uparrow$ : Higher is better.} & \multicolumn{2}{l}{$\downarrow$ : Lower is better.} & \multicolumn{4}{l}{$\# / \mathrm{km}$ : number of infractions per kilometer.} \\

\end{tabular}
\label{table-cadl}
\end{center}
\end{table*}

\section{CONCLUSION AND FUTURE WORKS}

In this paper, we propose the trajectory planning method for autonomous driving in urban environments and our solutions to conquer the CADL challenge. The trajectory generator produces a reliable and safe maneuver over the planning horizon. However, in several complex cases, the planner is not able to compose the most optimal path due to the fact that the parameters of the cost function are fixed during different scenarios. 

To solve this circumstance, adaptive cost functions could be added to the current approach, which requires an extensive review of the scenarios in the challenge. Furthermore, adding a behavior layer would fill the gap in the decision-making problem in our current approach. By applying the Partially observable Markov decision process and its alternatives, we can further increase the reasoning capability of our agent.

\end{document}